\documentclass[sigconf]{acmart}

\AtBeginDocument{%
  \providecommand\BibTeX{{%
    \normalfont B\kern-0.5em{\scshape i\kern-0.25em b}\kern-0.8em\TeX}}}

\usepackage{gensymb}
\begin{document}

\settopmatter{printacmref=false} 
\renewcommand\footnotetextcopyrightpermission[1]{} 
\pagestyle{plain} 

\title{An Efficient Point of Gaze Estimator for Low-Resolution Imaging Systems Using Extracted Ocular Features Based Neural Architecture}

\author{Atul Sahay}
\affiliation{%
  \institution{Indian Institute of Technology}
  \city{Bombay}
  \state{Maharashtra, India}}
  \email{atulsahay@cse.iitb.ac.in}

\author{Imon Mukherjee}
\affiliation{%
  \institution{Indian Institute of Information Technology}
  \city{Kalyani}
  \state{West Bengal, India}}
\email{imon@iiitkalyani.ac.in}

\author{Kavi Arya}
\affiliation{%
  \institution{Indian Institute of Technology}
  \city{Bombay}
  \state{Maharashtra, India}}
  \email{kavi@iitb.ac.in}






\begin{abstract}
 A user's eyes provide means for Human Computer Interaction (HCI) research as an important modal. The time to time scientific explorations of the eye has already seen an upsurge of the benefits in HCI applications from gaze estimation to the measure of attentiveness of a user looking at a screen for a given time period. The eye-tracking system as an assisting, interactive tool can be incorporated by physically disabled individuals, fitted best for those who have eyes as only a limited set of communication. The threefold objective of this paper is - 1.To introduce a neural-network-based architecture to predict users' gaze at 9 positions displayed in the $11.31\degree$ visual range on the screen, through a low resolution based system such as a webcam in real-time by learning various aspects of eyes as an ocular feature set; 2.A collection of coarsely supervised feature set obtained in real-time which is also validated through the user case study presented in the paper for 21 individuals ( 17 men and 4 women) from whom a 35k set of instances was derived with an accuracy\_score of 82.36\% and f1\_score of 82.2\%; and 3.A detailed study over applicability and underlying challenges of such systems. 
 
The experimental results verify the feasibility and validity of the proposed eye-gaze tracking model.  
\end{abstract}




\maketitle

\section{Introduction}
Users' gaze tracking is a means of analyzing an user's eye movements, recorded to estimate their focus or absolute point  of  gaze(POG)  on the computer's screen for a particular time instance.  Focus of attention plays an important role in user modeling and has significant application throughout the computer vision domain. The early development in the field of eye gaze tracking dated back to the 1879, when Louis Emile Javal had pointed out in his observations, the saccadic movement of eye instead of previously assumed sweep movement of an eye \cite{Huey} . Edmund's centuries-old experiment with aluminum pointer over the contact lens gave birth to the very first eye gaze tracker \cite{Huey}. However, researchers began the discovery in HCI, only when a very non-intrusive eye gaze tracking model was developed by Thomas Busewell in Chicago to track users' attention, by studying the properties of laser beam reflections on the eye. \cite{Bussel}.

In recent years, numbers of gaze tracking models have already been reported.  From the past few decades, mechanisms and dynamics of rotations of the eyes have been the area of focus for studies such as Electro-Oculography \cite{Kaufman}, pupil, eyelid trailing \cite{Ebisawa, Hutchinson}. However, looking at things from the perspective of an user, features drawn by a eye should be given more importance. Visual perception of a user can be studied through the saccadic movement of the eye. Many  such visualization techniques that work on the various features such as ocular characteristics through fixating blobs over eyes and heat map are reported in the literature \cite{Duchowski, Burch}.

The potential usage of gaze tracking in HCI with the proven efficiency \cite{Jacob} can be visualized with its direct applications not only lying in modal's input but even in the measurement of strain in employees of various sectors (from IT firms to Banking sector) \cite{MATJAZ}, a user's focal point over a screen is always the best input for the field of marketing, another unconventional use of technology can be seen via analysis of eye movement records of younger and elderly people as later makes more use of foveal vision than former. \cite{Fukuda}.

A current trend in eye-tracking has been shown through various analysis reported in \cite{Bulling}, research in the domain has shifted from laboratory settings to various mobile settings of different indoor and outdoor environments. Due to upsurge in accessibility and sophistication of POG trackers, applicability in the commercial sector has been increased too. Applications of which can be seen as the web usability, advertising, sponsorship, package design and automotive engineering \cite{Net}. However; on the conventional side, POG as a modal input can also be used as a direct controller of graphical user interfaces (GUI), which is proven to be more efficient (like the MAGIC system \cite{Zhai}) than the traditional input devices such as a mouse.

While developing a POG tracker, bifurcation is done based on the level of physical intrusion employed- active methods and passive methods. Active method has the requirement of special hardware to increase the accuracy, e.g., infrared cameras and illuminators, making it expensive and invasive \cite{Bergasa}. Whereas the passive method employs the use of non invasive technique such as use of a standard remote camera only. A current eye gaze tracker relies on intrusive technique such as measuring the corneal reflections through some light (usually infrared light) that is shone onto the eyes \cite{Guestrin}, measuring the electric potential around the eyes or using a special contact lens that facilitates the gaze tracking \cite{Glenstrup}. However, due to a strong imposition for controlled environment requirements such as motion dynamics or illumination makes the setup computationally sensitive and therefore shadowing their remarkable real-time performance.

\begin{figure}
\centerline{\includegraphics[width=8cm, height=4cm]{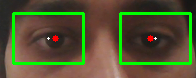}}
\vspace{0.3cm}
\centerline{\includegraphics[width=8cm, height=4cm]{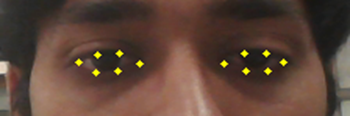}}
\caption{a) Region of Interest (ROI); pupil-center
\\
 b) detected landmarks over ROI}
\label{figure1}
\end{figure}

An alternative passive approach for users' gaze tracking has been discussed in \cite{Sahay}, where a low resolution camera based module is used to capture the real-time facial landmarks which are then used further in order to capture precisely the characteristic features from the region of an eye, ``Fig.~\ref{figure1}''.
Furthermore, these characteristic features are then used to devise an interesting ocular feature set, which is used later with simple heuristic of estimation for determining of an users' POG. 
However,  simple heuristics of estimation has never been sufficient enough in getting perspicacity of ocular's perception rich feature set, which allows more precise models to be drawn if more insights can be learned through these ocular parameters.

 In this paper, we have proposed a Feed-Forward Neural Network model based approach; which upon feeding extracted ocular perception rich feature set from a low resolution based input module, can detect users' POG in real-time. The preparation of such granular supervised feature sets along with the architecture is also discussed in line.

Main contributions of this paper are:

1. A non-invasive novel Feed-Forward Neural Network based architecture is proposed which can detect users' POG in real time and in varying mobile settings. 

2. A real-time coarse granularity, supervised data-set technique is also discussed. 

3. Quantitative and qualitative results are presented to measure accuracy and feasibility of the proposed algorithm.

4. The effect of selection of features for the training is properly discussed along with the domains where it can be applied..

The rest of the paper is structured as follows: The proposed architecture and ocular feature set is presented in section 2., experimental validation and evaluation is presented in section 3., a set of applications where such a low-resolution based eye gaze tracker can be utilized to operate a GUI and more are detailed in section 4. and finally in section 5. we have drawn our concluding remarks.

\section{Proposed method}
The entire architecture's proposed by us is based on the learning insights from extracted ocular features of the detected eye region such as iris displacement statistics with respect to a given reference point and eyelid opening statistics through eye aspect ratio (EAR) \cite{Tereza}. For robustness over the indoor and outdoor mobile settings, we have made use of an in-the-wild dataset collected from 21 individuals ( 17 men and 4 women )  with a total of 35,154 instances. A final prediction of user's POG is done by a shallow fully connected Feed-Forward Neural Network over 9 directions taken on the screen(north-west, north, north-east, east, south-east, south, south-west, west, center) "Fig. \ref{figure5}" .

\subsection{Eye Region Detection}
To extract the region of interest(ROI), we first made use of the Viola-Jones based object detector \cite{Viola} to get the facial dimensions from a low resolution camera based module. We employed a decade old Viola-Jones type object detector in our work only because of the proficiency it gives on low-resolution images in real-time especially on low computational devices.  

To find out the dimensions of the face, we need to keep in check the orientation of head, as the framework is sensitive towards face-camera tilt, which can produce  erroneous results when tilted too much. 
Typically, frameworks associated with object detection include features of rectangular white-black strips resembling haar basis features to note facial features. A cascade of trained models is then trained on these characteristics. A comparative increase in the power of the classifier through cascading can be seen from low to high stages. Grouping of features is done along the axis of the stages, with as many as hundreds of features per layer being classified. AdaBoost, a learning algorithm, is used to interface two layers so that only the best features can be learned.

\begin{figure}
\centerline{\includegraphics[width=9cm, height=9cm]{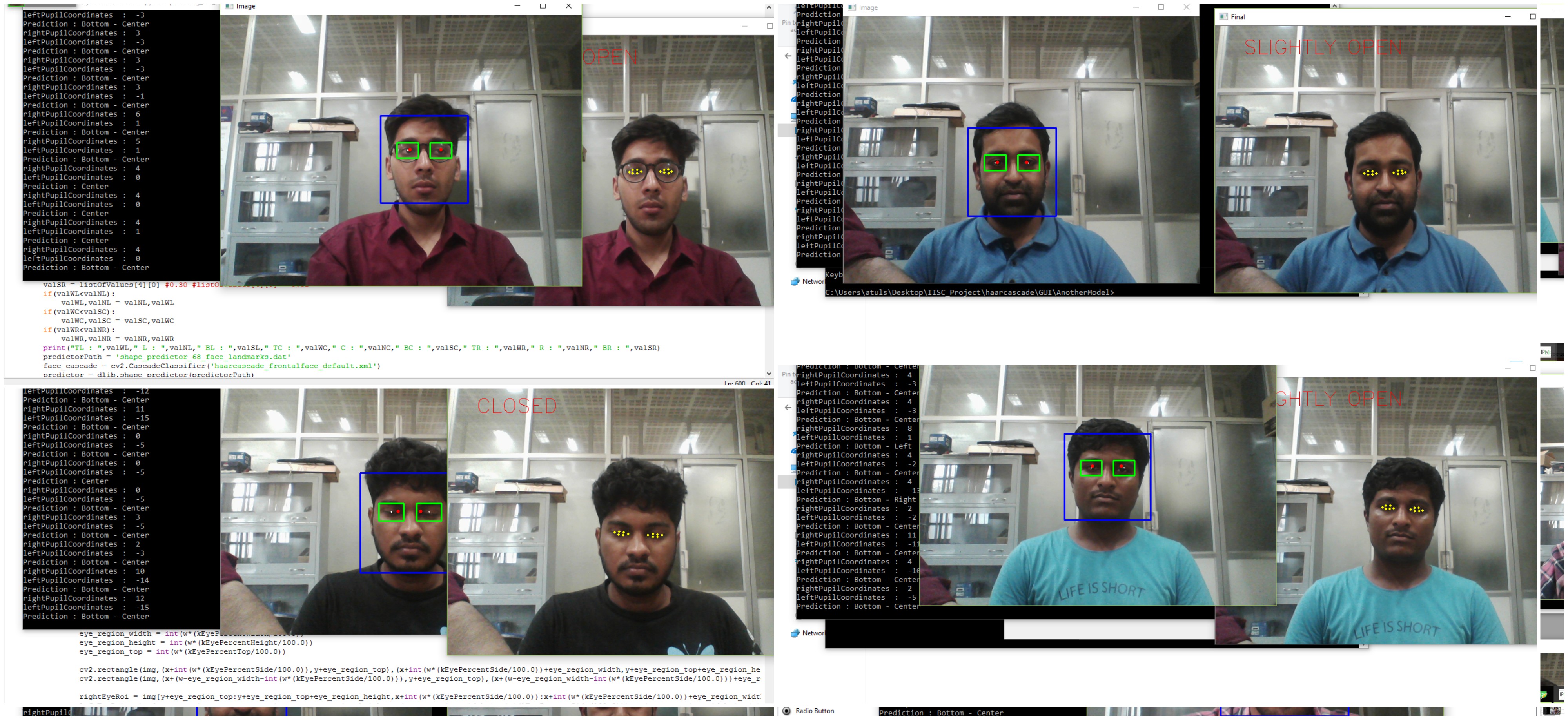}}
\vspace{0.3cm}
\centerline{\includegraphics[width=2cm, height=2cm]{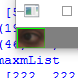}}
\caption{a) The extracted eye region(shown in green rectangle) over the facial dimension 
\\*
 b) The eye region after its being scaled down while maintaining the pixel density.
}
\label{figure2}
\end{figure}

From our extensive analysis, we have found that using straight haar features to extract the eye region is computationally more expensive than extracting it from the facial dimensions because we noted that the eye area always lies on the face with approximately fixed proportions ``Fig.~\ref{figure2}''. Once ROI is extracted we further scaled it down by maintaining the pixel density for fast processing while putting precision-computational complexity trade-off at stake.
 
Scaling down is an adaptable process that scales down the ROI based on a varying lightning condition (70-200 lux).

\subsection{Localization of iris-center}
The localization of the pupil-center plays a significant role in determining the ocular feature set. For high resolution camera modules, this can be easily done through the eclipse fitting \cite{Fabian},or dilation techniques \cite{Guestrin}.  To preserve the general applicability of the architecture, we tried these techniques with our low-resolution images captured by the standard webcam but found the trade-off between the precision and real-time application of such techniques. 

Using the knowledge that the pupil-center or iris center is the darkest area in the eye, we have devised a kernel that we convolved over the ROI for fixating the pupil-center. 

Besides this, eyelashes and eyelid corners are raised in the same proportion as the pupil-center in the inverse grayscale image``Fig.~\ref{figure4}'', to overcome this we clipped off the pixels, connected to the image border.

\begin{figure}
\centerline{\includegraphics[width=9cm, height=7cm]{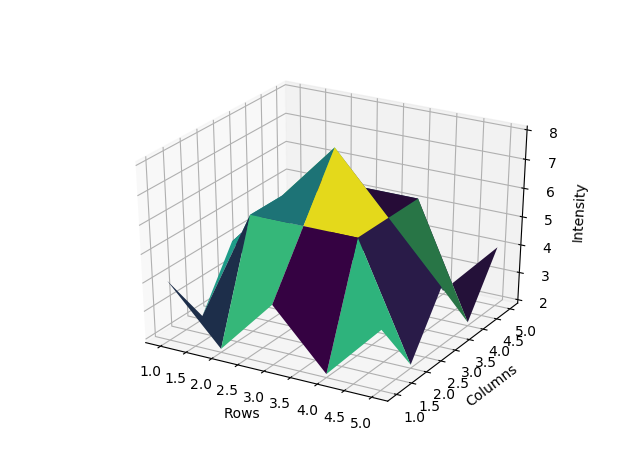}}
\vspace{0.3cm}
\caption{ 3D heat-map of filter(5x5) used for localizing the iris-center.
}
\label{figure3}
\end{figure}

While fixating the pupil-center, we prioritize more on evaluating several candidates, instead of a single maximum point selection as the darkest region, a number of candidates are evaluated is based on the presence of the lightning condition i.e. for sufficient lighting, the number of candidates is less and vice-versa. Information of the neighboring pixels is used to assign weight score to each candidate, for this convolution the kernel used is shown in ``Fig.~\ref{figure3}'' for making the best choice among them as the pupil-center.


\subsection{Eye Aspect Ratio}
Eyelid opening as an ocular parameter has been widely used in literature eg. for driver attentiveness, drowsiness detection, stress measuring techniques, but never exploited much for eye gaze tracking. However, eye gaze tracking has been worked on only by author \cite{Sahay} where the users' gaze estimation is done easily by using eye aspect ratio (EAR)\cite{Tereza}.

\begin{figure}
\centerline{\includegraphics[width=3cm, height=2cm]{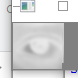}}
\vspace{0.3cm}
\centerline{\includegraphics[width=3cm, height=2cm]{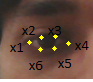}}
\caption{a) The inverse grayscale of ROI
\\*
b) 2D landmarks constructed on eyelids.
}
\label{figure4}
\end{figure}

Using the landmark formation over the eye region, a scalar quantity representing the geometric aspect of eyelid is extracted.\cite{Tereza}.
\begin{equation}
EAR=\frac{(||x2-x6||+||x3-x5||)}{(2*||x1-x4||)} 
\end{equation}
Over the eye, landmark formation - x1$\ldots$x6 are depicted in 
''Fig.~\ref{figure4}''.

Eyelid opening can be easily measured through EAR, which is greater for large contour formation by eyelids and vice versa. This involuntary enhancement of the eyelid contour becomes the basis of the ocular feature set; when one tries to view the material placed on top of the screen. The EAR remains nearly constant, with varying distances between the user and the camera module, making it unchanged for minor changes in distance.


\begin{table*}
\caption{Ocular Feature Set}
\begin{center}
\centering
 \begin{tabular}{||c c c c c c c||} 
 
 \hline
 Aspect Ratio & minR & maxR & Displacement & minD & maxD & Class \\
 \hline\hline
 0.36 &	0.277538696	& 0.430383655 &	3.08 & -2 &	11 & 1\\
 0.35 &	0.316103573	& 0.462965465 &	2.76 & -7 &	10 & 2\\
 0.36 &	0.312934426	& 0.436084602 & -1.62 &	-8 & 7 & 3\\
 0.31 &	0.276095344	& 0.391149739 &	-2.82 &	-8 & 7 & 4\\
 0.27 &	0.243460121	& 0.338916135 & -5.2  &	-10 & 1 & 5\\
 0.27 &	0.221078406	& 0.296966443 &	0.68 &	-5	& 3 & 6\\
 0.26 &	0.217213378	& 0.297732295 &	9.68 &	1 &	13 & 7\\
 0.28 &	0.24662586	& 0.348253656 & 8.74 & -5 &	12 & 8\\
 0.32 &	0.248078475 & 0.398217235 &	1.1	 & -7 &	9 & 9\\
 \hline
 \end{tabular}%
\label{table1}
\end{center}
\footnotesize{ \textsuperscript{minR}minimum EAR , \textsuperscript{maxR}maximum EAR , \textsuperscript{minD}minimum Displacement , \textsuperscript{maxD}maximum Displacement }\\
\end{table*}

\subsection{Ocular Feature Set}

After the considerable extraction of eye region from the facial dimensions of the users, extracted ocular characteristics forms the key features. The measure of displacement of the iris-center from the reference point defines the relative shift of gaze of the user in the horizontal axis and EAR defines the relative shift of gaze over the vertical axis. For better understanding of visual perception, offsets are also added to the feature set which marks the error range of users' POG.

In the table \ref{table1}, column heads: Aspect Ratio, minR, maxR, Displacement, minD, maxD, class; depicts the EAR, minimum value of EAR noticed for that particular class, maximum value of EAR noticed for the class, displacement measure of the iris-center from the reference point, minimum displacement observed , maximum displacement observed and class labels 1$\ldots$9  over 9 possible directions (north-west, north, north-east, east, south-east, south, south-west, west, center) respectively. 

These 9 directions provide coarse supervision to the network indicating where the users' POG has been shifted. We created our calibration suite shown in  "Fig.\ref{figure5}"  that provides the class label information to the particular gaze instance in real-time. We have collected an in-the-wild dataset from 21 individuals (17 men and 4 women) with a total of 35,154 instances shown in ''Table \ref{table1}''. We have avoided controlled experimental settings as much as possible to make the dataset stronger.

\subsection{Feed Forward Network}
The final section of the architecture contains a shallow fully connected Feed-Forward Neural Network, which is applied to each POG instance separately and identically. This consists of 4 linear transformations with a ReLU activation in between.

\begin{equation}
\texttt{FFN} = W_4\texttt{RELU}(W_3\texttt{RELU}(W_2\texttt{RELU}(W_1x + b_1) + b_2) + b_3) + b_4
\end{equation}
\begin{equation}
\texttt{RELU(Z) = max(0,Z)}    
\end{equation}

Where $W_1, W_2, W_3, W_4$ are the parameter matrices associated with each hidden layer of the FFN; while $b_1, b_2, b_3, b_4$ are the biases associated with each hidden layer and "x" is the input ocular feature vector.

To include better predictors with higher confidence in a multi-class classification setting, we used the negative log in this function because it showed interesting behavior which is that for a low confidence score for a class, it tends to infinite loss; thus leading to greater punishment and vice versa. For $\widehat{y}$ predicted value, training loss over n samples is defined as: 
\begin{equation}
    \texttt{Loss} = -\frac{1}{n}\sum_{i=1}^{n}\log \left ( \widehat{y}^{(i)} \right )
\end{equation}

Softmax functionality is widely used with the likelihood loss for obtaining the confidence score relative to each other class. We made use of the log softmax functionality; even though both these functions show monotonicity, their effect on the relative values of the loss function changes, log-softmax will punish bigger mistakes in likelihood space, more. For prediction scores of $f^{1} ,\ldots  ,f^{k}$ for each class $1 ,\ldots ,k$, the log softmax functionality is defined as follows:

\begin{equation}
    \texttt{softmax}  = \frac{\exp {f^{k}}}{\sum_{i=1}^{K}\exp {f^{i}}}
\end{equation}

\begin{equation}
    \texttt{log softmax} = \log \left ( \exp {f^{k}} \right ) - \log \left ( {\sum_{i=1}^{K}\exp {f^{i}}} \right )
\end{equation}

Second reason of choosing the latter functionality could be seen by the numerical stability it provides in case of the numerical underflow or overflow scenarios. 

The final prediction is the class in which the network has the most confidence.
 
\section{Evaluation and Results}
For accurately bench marking the feasibility and reliability of the architecture, we have constructed our test module depicted in  ''Fig.~\ref{figure5}'' that precisely measures the following scores: accuracy\_score, precision, recall, f1\_score. To do so we had asked 21 individuals ( 17 men and 4 women, the average age of this group being 21.33  years ) to participate in the evaluation suite.

\subsection{Environment settings:}
To maintain the breadth of architecture, we ensured that we do not restrict to  a controlled environment, so we used the following environment for our user studies:

\begin{itemize}
    \item   Lightning setting: 80-200 lux
    \item	Webcam : 0.9 megapixels(HD) @ 30 fps
    \item	Processor: Intel Core i5-5200U 5th Gen Processo @2.20GHz, turbo boost to 2.70GHz
    \item   RAM : 8GB DDR3
    \item	Display : 15.6-inch LED Backlit Display (1368X768)
\end{itemize}

\subsection{Evaluation Suite}
\begin{figure}
\centerline{\includegraphics[width=9cm, height=9cm]{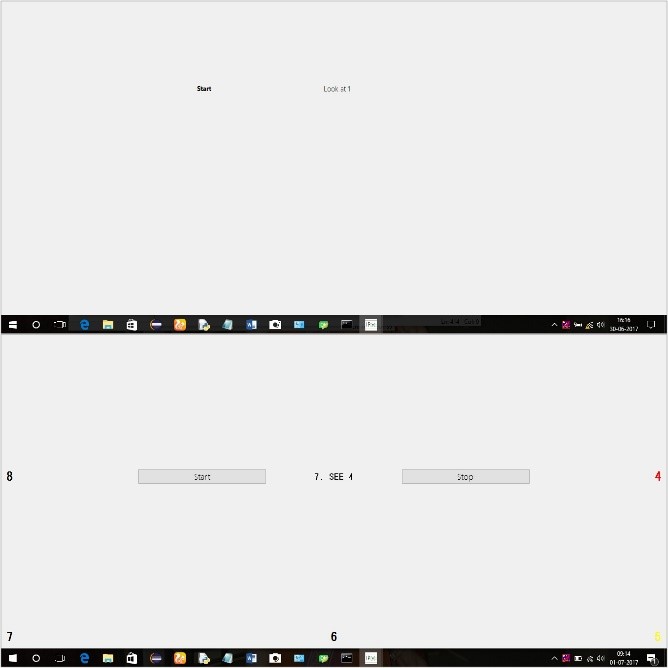}}
\caption{Evaluation suite.
}
\label{figure5}
\end{figure}

The evaluation suite is designed keeping in mind the application of the architecture, 9 class labels representing 9 directions are placed on the screen with a visual angle within 11.31$\degree$ 'Fig.~\ref{figure5}''. The design was kept simple so as not to produce any cognitive load.

Every user undergoing the evaluation user metric was given simple instructions not to make a big head movement, because ocular features are largely sensitive to head-camera tilt, thus providing an incorrect evaluation. The evaluation suite can occupy 50 instances of ocular features for one second of the class label shown.

\subsection{Evaluation Metrics}
The dataset on which the FFN was trained is sufficient with over 35k training instances, as a rule of thumb, we went with 3780 test instances marking the 90:10 training and test set size ratio.

We made use of the accuracy function and $f1_{score}$ as our evaluation metrics.

The accuracy function is computed on the basis of number of correct predictions made by the model. For every predicted class label $\hat{y}$ and ground truth y, accuracy-score for $n_{samples}$ test instances is defined as follows:

\begin{equation}
    \texttt{accuracy}(y, \hat{y}) = \frac{1}{n_\text{samples}} \sum_{i=0}^{n_\text{samples}-1} 1(\hat{y}_i = y_i)
    \label{eq7}
\end{equation}
\begin{equation}
    \texttt{1(x=y)} =\left\{\begin{matrix} 
1, \texttt{if y==x}
\\
0, \texttt{if y!=x}
\end{matrix}\right.
\label{eq8}
\end{equation}

Since the applicability of the architecture depends largely on the precision and recall of the class labels, we also included f1\_score in our evaluation metric.

\begin{equation}
    \text{precision} = \frac{tp}{tp + fp},
    \label{eq9}
\end{equation}
\begin{equation}
    \text{recall} = \frac{tp}{tp + fn},
    \label{eq10}
\end{equation}
\begin{equation}
    F = 2 \frac{\text{precision} \times \text{recall}}{ \text{precision} + \text{recall}}.
    \label{eq11}
\end{equation}
where tp, fp, fn are true positive, false positive and false negative respectively.

\subsection{Result}

\begin{figure}
\centerline{\includegraphics[width=9cm, height=9cm]{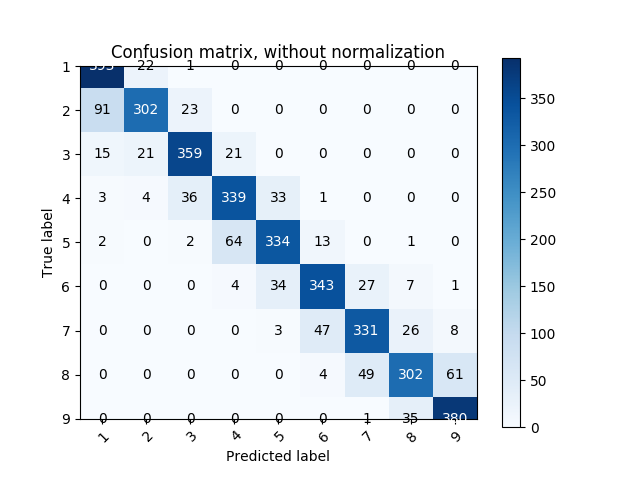}}

\centerline{\includegraphics[width=9cm, height=7cm]{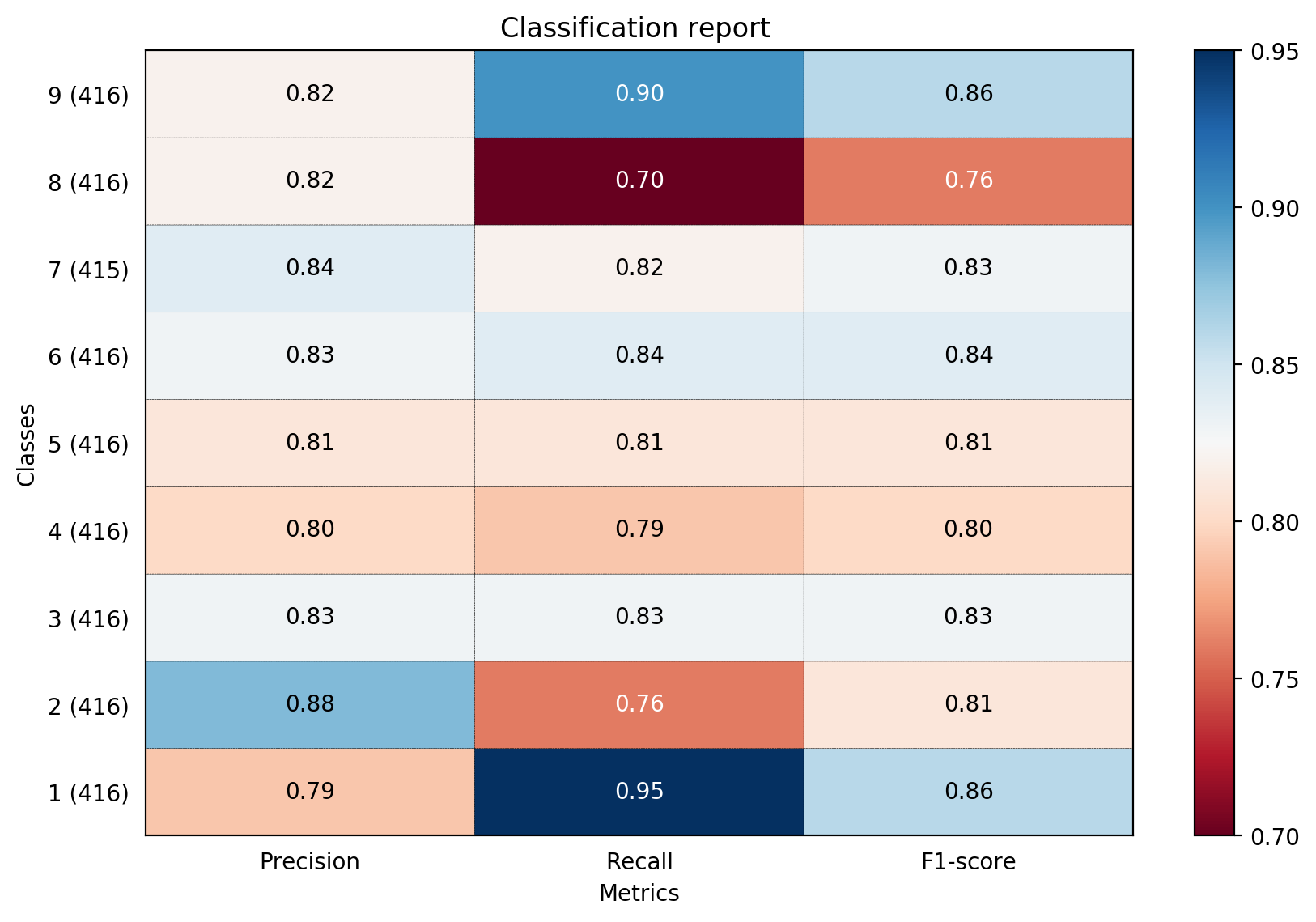}}
\caption{a) A 3D heat map of the confusion matrix
\\*
b) Heap map of the classification - report for visualization.
}
\label{figure6}
\end{figure}

We reported accuracy\_score (see eqs [\ref{eq7}, \ref{eq8}]) of 82.36\% and f1\_score (see eqs [\ref{eq9}, \ref{eq10}, \ref{eq11}]) of 0.82 (shown in table \ref{table2}) which is comparatively good for low resolution based cameras. As we have used a balanced test set hence the macro averaging and weighted(micro) averaging comes equal. For better visualization a detailed per class label recall and precision values are depicted in ''Fig \ref{figure6}b''; while ''Fig \ref{figure6}a'' shows the confusion that happened between the class label predictions in the form of a confusion matrix.

\begin{table}[h]
\centering
\caption{Performance on test set }
\begin{tabular}{|l|l|l|l|l|l|}
\hline
Scoring-Type & Precision & Recall & F1-score & Score & Support\\ \hline
accuracy & &  & & 0.82 & 3780 \\ \hline
macro avg & 0.83 & 0.82 & 0.82 & & 3780 \\ \hline
weighted avg & 0.83 & 0.82 & 0.82 & & 3780 \\ \hline
\end{tabular}
\label{table2}
\end{table}

Support (number of occurrences of each class in ground truth) is shown in the brackets in the classification report ''Fig.~\ref{figure6}''

\section{Discussion And Application}
Despite using shallow FFN at a low-resolution camera-input our test model performed well on the test set. With an accuracy of 82.36\%, the proposed architecture can distinguish between nine locations on the screen (eight locations on the edge and one on the middle of the screen) within a radius of 11.31$\degree$ of the viewing angle.

For a more in-depth analysis of the characteristic behavior of ocular features, we formulated a confusion matrix and found that the model is confused a lot between point label 1 and 2 while predicting position 2, a similar discrepancy is also shown by the point label pairs 8 and 9. One possible explanation of such disparity is the user distraction while undergoing the evaluation suite but we would like to investigate more on this before making any conclusions.

Furthermore, instead of always making per-frame predictions, if we accumulate knowledge of some frames; the model accuracy easily surpasses reported accuracy and in few cases, it increases to 94\%; we have not yet reported it, as we left it for future investigation.

Unquestionably, the accuracy is not comparable to the state-of-the-art infrared gaze trackers; but the presented model is a non-intrusive eye-tracking and gaze monitoring system, which tracks and locate the users' pupil as soon as user appears in the view of a camera, without any special lightning and special marks on the face. The only requirement of this passive model is of low resolution cameras; thus making it reliable as the commercially available eye-gaze tracker. The data set required for training the model is an in-the-wild data set that can be easily collected from the user using the model without any prior involvement of setup for making the data set. 

The proposed model accuracy surpasses the one reported in the literature\cite{DostalJ.}, where users' POG are used to select a monitor view in multiple monitor setups.

Through our extensive case study for eye-to-control systems as input modals, we identified a set of applications where the proposed system may be of use:
\begin{enumerate}
    \item The proposed model may work for the case where a modal controlled system having a limited number of selection buttons, but is difficult to navigate using conventional interactions. for example; The fire panel in the hand of the firefighter shows the defective area, intensity of heat or smoke in different parts of the place, where the traditional touchscreen will not be of much help due to thick gloves and sweat.
    
    \item Nowadays, due to the increase in the number of users favoring mobile devices for watching sitcoms, as most video content is not made keeping the screen dimension in mind, the projection of video content to such smaller screens will provide a video on which the central objects may become indistinguishable from the rest. One solution is to focus on the most visually interesting parts of the video and then reproduce the video content appropriately. As reported by a study \cite{chamaret} of more than 16 individuals, low-cost users' gaze tracker has performed surprisingly well, showing that more than 90\% of the focus areas are placed in reframed clips. The proposed system has control over 9 directions and as such it is useful for creating a bounding in the region of interest, where the video is focused to remodel.
    
    
    \item	The system can be serviced as an assisting tool for people with severe speech and motor impairments (SSMI), limited with eye as the prime modal.    As per detailing in a user case study \cite{atul} of four students with SSMI studying at the Spastic Society, The GUI can be designed which may contain at least 9 buttons ''Fig.\ref{figure8}'' which can identify the selection in not more than 2 seconds. These students were able to correctly select 35 out of 40 questions in an online quiz application using web-based Gaze Tracker.
    
    \begin{figure}
    \centerline{\includegraphics[width=4cm, height=4cm]{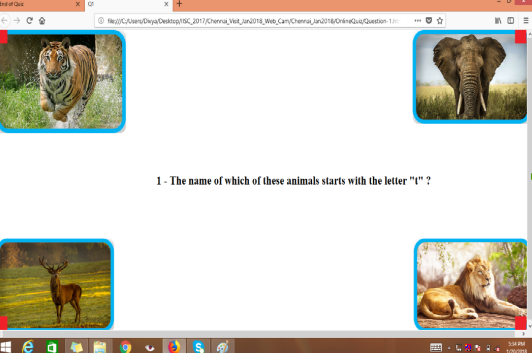}}
    \caption{a) GUI model used for study of SSMI students. 
    }
    \label{figure8}
    \end{figure}

    \item The system can also be used as a reliable real-time input modal for HCI, which is best for physically disabled people with limited modality
.To aid the use of an user's gaze tracking for human-computer communication, a state-of-the-art system has been discussed in the literature\cite{zhang2017eye}, where a virtual mouse interface based on users' gaze is used for navigation along with the integration of both mouse and keyboard function. Furthermore, with the tweak of the customization, the proposed system can also serve as an additional modality in multi-modal systems where it can be used for zooming in and out of part of the display [24], [25]

\end{enumerate}


\section{Conclusion}
Through our proposal, we have shown how learning a few insights from ocular features can largely make a difference in real-time tracking of users' POG. Such models are feasible, computationally inexpensive and also reliable.
The training of the model on the in-the-wild data set makes the system versatile in contrast to the commercially available eye-gaze tracker that requires a large data set for training in a very controlled environment.
With some adaptive settings encapsulated by the model, it is superior to some existing gaze monitoring devices that are sensitive to the controlled environment settings. 
We have demonstrated the applicability of the system as an adjunct to modal controlled devices which expands the width of its use. Whether Firefighter's hands-on mounted device, a virtual modal interface for reliable navigation or an assisting interactive tool for SSMI people. In the current model, deriving the users' focus of attention from low eye gaze patterns has not been addressed, to do so a high dimensional model is needed that can also incorporate the 3D map of users' sight view. Further involuntary eye movements fed significant distortions in the training data set, that eventually decreases the accuracy.

\bibliographystyle{ACM-Reference-Format}
\bibliography{ref}

\end{document}